\def\year{2018}\relax
\documentclass[letterpaper]{article} 
\usepackage{aaai18}  
\usepackage{times}  
\usepackage{helvet}  
\usepackage{courier}  
\usepackage{url}  
\usepackage{graphicx}  
\usepackage{arydshln}
\begin{document}
%
\title{Modular Architecture for StarCraft II with Deep Reinforcement Learning}
\author{Dennis Lee*, Haoran Tang*, Jeffrey O Zhang, Huazhe Xu, Trevor Darrell, Pieter Abbeel\\
University of California, Berkeley\\
\{dennisl88, hrtang.alex, jozhang, huazhe\_xu\}@berkeley.edu, \{trevor, pabbeel\}@eecs.berkeley.edu \\
(* Equal contributions)
}

\newcommand{\namecite}[1]{\citeauthor{#1} (\citeyear{#1})}
\newcommand{\pmm}{$\pm$}

\maketitle
\begin{abstract}
We present a novel modular architecture for StarCraft II AI. The architecture splits responsibilities between multiple modules that each control one aspect of the game, such as build-order selection or tactics. A centralized scheduler reviews macros suggested by all modules and decides their order of execution. An updater keeps track of environment changes and instantiates macros into series of executable actions. Modules in this framework can be optimized independently or jointly via human design, planning, or reinforcement learning. We apply deep reinforcement learning techniques to training two out of six modules of a modular agent with self-play, achieving 94\% or 87\% win rates against the "Harder" (level 5) built-in Blizzard bot in Zerg vs. Zerg matches, with or without fog-of-war.
\end{abstract}

\begin{table*}[t]  
\centering
\begin{tabular}{|l|l|l|l|} 
 \hline
 Module             & Responsibility                                                        & Current Design \\ [0.5ex] 
 \hline
 Worker management  & Ensure that resources are gathered at maximum efficiency                & Scripted     \\
 Build order        & Choose what unit/building/upgrade to produce                            & FC policy    \\
 Tactics            & Choose where to send the army (attack or retreat)                       & FCN policy   \\
 Micromanagement    & Micro-manage units to destroy more opposing units  & Scripted     \\
 Scouting           & Send scouts and track opponent information    & Scripted + LSTM prediction    \\
 \hline
\end{tabular}
\caption{The responsibility of each module and its design in our current version. FC = fully connected network. FCN = fully convolutional network.}
\label{table:modules}
\end{table*}

\section{Introduction}

\noindent Deep reinforcement learning (deep RL) has become a promising tool for acquiring competitive game-playing agents, achieving success on Atari \cite{mnih_2015_atari}, Go \cite{silver_2016_alphago}, Minecraft \cite{tessler_2017_minecraft}, Dota 2 \cite{openai_2018_dota_five},
and many other games. It is capable of processing complex sensory inputs, leveraging massive training data, and bootstrapping performance without human knowledge via self-play \cite{silver_2017_alphazero}. However, StarCraft II, a well recognized new milestone for AI research, continues to present a grand challenge to deep RL due to its complex visual input, large action space, imperfect information, and long horizon. In fact, the direct end-to-end learning approach cannot even defeat the easiest built-in AI \cite{vinyals_2017_starcraft}. 

StarCraft II is a real-time strategy game that involves collecting resources, building production facilities, researching technologies, and managing armies to defeat the opponent. Its predecessor StarCraft has attracted numerous research efforts, including hierarchical planning \cite{weber_2010_gda} and tree search \cite{uriarte_2016_mcts} (see survey by \namecite{ontanon_2013_survey}). Most prior approaches focus on substantial manual designs, yet still unable to defeat professional players, potentially due to their inability to utilize game play experiences \cite{mit_2017_starcraft}. 

\begin{figure}
    \centering
    \includegraphics[width=0.45\textwidth]{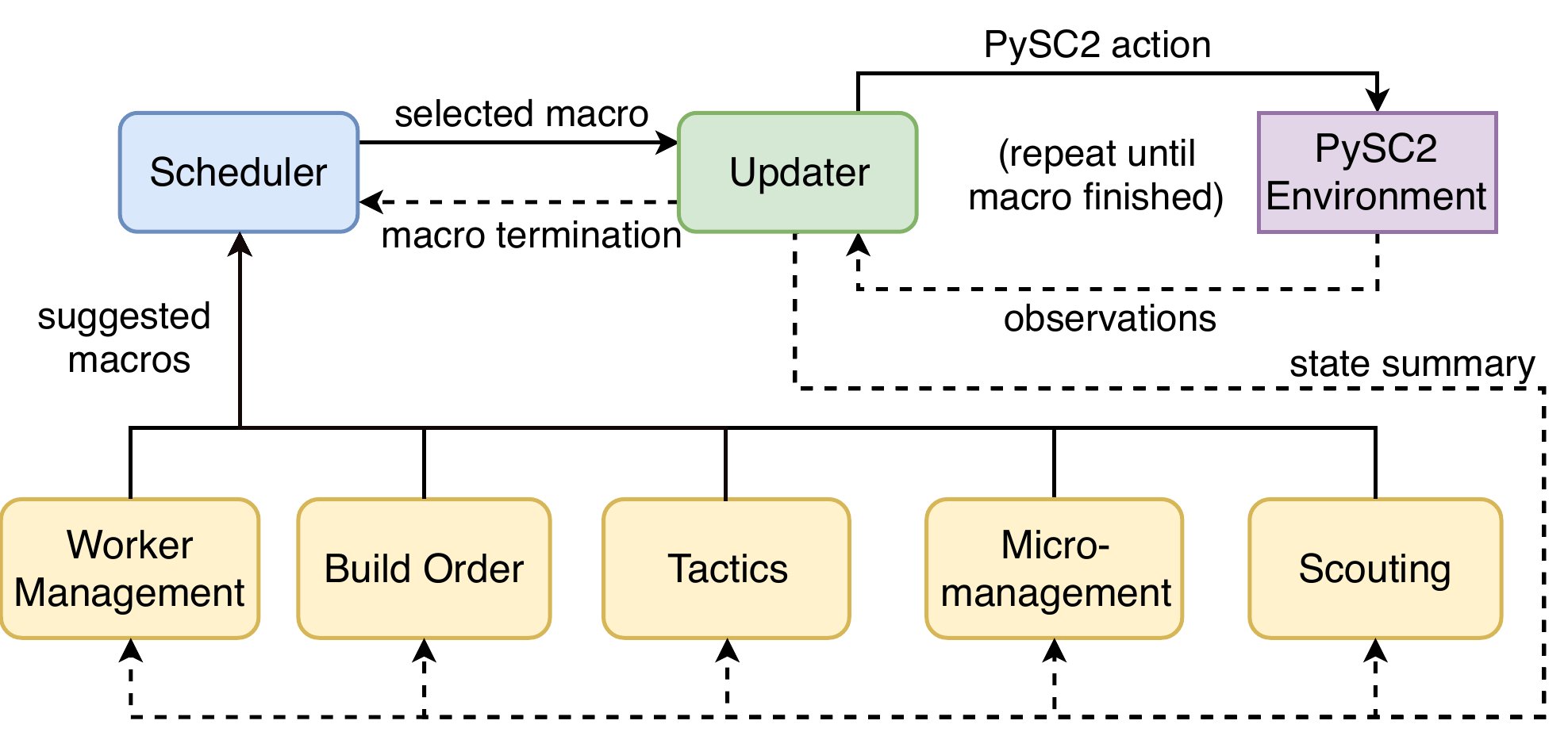}
    \caption{The proposed modular architecture for StarCraft II}
    \label{fig:modular_architecture}
\end{figure}
We believe that deep RL with properly integrated human knowledge can effectively reduce the complexity of the problem without compromising policy expressiveness or performance. To achieve this goal, we propose a flexible modular architecture that shares the decision responsibilities among multiple independent modules, including worker management, build order, tactics, micromanagement, and scouting (Figure~\ref{fig:modular_architecture}). Each module can be manually scripted or handled by a neural network policy, depending on whether the task is routine and hence easy to handcraft, or highly complex and requires learning from data. All modules suggest macros (predefined action sequences) to the scheduler, which decides their order of execution. In addition, an updater keeps track of environment information and adaptively executes macros selected by the scheduler.

We further evaluate the modular architecture by reinforcement learning with self-play, focusing on important aspects of the game that can benefit more from massive training experiences, including build order and tactics. The agent is trained on the PySC2 environment \cite{vinyals_2017_starcraft} that exposes a challenging human-like control interface. We adopt an iterative training approach that first trains one module while others follow very simple scripted behaviors, and then replace the scripted component of another module with a neural network policy, which continues to train while the previously trained modules remain fixed. We evaluate our agent playing Zerg v.s. Zerg against built-in bots on ladder maps, obtaining win rates 94\% or 87\% against the ``Harder'' bot, with or without fog-of-war. Furthermore, our agent generalizes well to held-out test maps and achieves similar performance. 

Our main contribution is to demonstrate that deep RL and self-play combined with the modular architecture and proper human knowledge can achieve competitive performance on StarCraft II. Though this paper focuses on StarCraft II, it is possible to generalize the presented techniques to other complex problems that are beyond the reach of the current end-to-end RL training paradigm.

\section{Related Work}
\noindent Classical approaches towards playing the full game of StarCraft are usually based on planning or search. Notable examples include case-based reasoning \cite{aha_2005_cbp}, goal-driven autonomy \cite{weber_2010_gda}, and Monte-Carlo tree search \cite{uriarte_2016_mcts}. Most other research efforts focus on a specific aspect of the decision hierarchy, namely strategy (macromanagement), tactics, and reactive control (micromanaging). See the survey of \namecite{ontanon_2013_survey} and \namecite{robertson_2014_review} for complete summaries. Our modular architecture is inspired by hierarchical and modular designs that won past AIIDE competitions, especially UAlbertaBot \cite{churchill_2017_ualbertabot}, but features the integration of deep RL training and self-play instead of intense hard-coding.

Reinforcement learning studies how to act optimally in a Markov Decision Process to maximize the discounted sum of rewards $R = \sum_{t=0}^T \gamma^t r_t \ (\gamma \in (0, 1])$. The book by \namecite{sutton_1998_reinforcement} gives a good overview. Deep reinforcement learning uses neural networks to represent the policy and/or the value function, which can approximate arbitrary functions and process complex inputs (e.g. visual information).

Recently, \namecite{vinyals_2017_starcraft} have released PySC2, a python interface for StarCraft II AI, and evaluated state-of-the-art deep RL methods. Their end-to-end training approach, although shows potential for integrating deep RL to RTS games, cannot beat the easiest built-in AI. Other efforts of applying deep learning or deep RL to StarCraft (I/II) include controlling multiple units in micromanagement scenarios \cite{peng_2017_multiagent,foerster_2017_counterfactual,usunier_2017_episodic,shao_2018_micro} and learning build orders from human replays \cite{justesen_2017_macro}. To our knowledge, no published deep RL approach has succeeded in playing the full game yet. 

Optimizing different modules can also be cast as a cooperative multi-agent learning problem. Apart from aforementioned multi-agent learning works on micromanagement, other promising methods include optimistic and hysteretic Q learning \cite{lauer_2000_algorithm,matignon_2007_hysteretic,omidshafiei_2017_multitask}, and centralized critic with decentralized actors \cite{lowe_2017_multi}. Here we use a simple iterative training approach that alternately optimizes a single module while keeping others fixed, though incorporating multi-agent learning methods is possible and can be future work. 

Self-play is a powerful technique to bootstrap from an initially random agent, without access to external data or existing agents. The combination of deep learning, planning, and self-play led to the well-known Go-playing agents AlphaGo \cite{silver_2016_alphago} and AlphaZero \cite{silver_2017_alphazero}. More recently, \namecite{bansal_2018_emergent} has extended self-play to asymmetric environments and learns complex behavior of simulated robots.

This work was developed concurrently with the hierarchical, modular design by \namecite{sun_2018_tstarbots}, which also utilizes similar macro actions. An important difference is that our agent is trained by playing against only itself and several scripted agents under the modular architecture, but never sees the built-in bots until the evaluation time.

\begin{table*}[t]  
\centering
\begin{tabular}{|l|l|l|} 
 \hline
 Module        & Macro name and inputs               & Executed sequence of macros or PySC2 actions \\ [0.5ex] 
 \hline
 (All)         & \textit{jump\_to\_base} (base)      & move\_camera (base.minimap\_location) \\
               & \textit{select\_all\_bases}         & select\_control\_group (bases\_hotkey)  \\
 Worker        & \textit{rally\_workers} (base)      & (1) \textit{jump\_to\_base} (base), (2) \textit{select\_all\_bases}, \\
               &                                     & \quad \quad (3) rally\_workers\_screen (base.minerals\_screen\_location) \\  & \textit{inject\_larva}              & (1) select\_control\_group (Queens\_hotkey), (2) for each base: \\
               &                                     & \quad \quad  (2.1) \textit{jump\_to\_base} (base),   \\
               &                                     & \quad \quad (2.2) effect\_inject\_larva\_screen (base.screen\_location) \\
 Build order   & \textit{hatch} (unit\_type)         & (choose by unit\_type, e.g. Zergling $\rightarrow$ train\_zergling\_quick)\\
               & \textit{hatch\_multiple\_units} (unit\_type, $n$) & (1) \textit{select\_all\_bases}, (2) select\_larva, (3) \textit{hatch} (unit\_type) $n$ times\\
               & \textit{build\_new\_base}           & (1) base =  closest unoccupied base (informed by the updater) \\
               &                                     & \quad \quad (2) \textit{jump\_to\_base} (base), (3) select\_any\_worker, \\ 
               &                                     & \quad \quad (4) build\_hatchery\_screen (base.screen\_location) \\
 Tactics       & \textit{attack\_location} (minimap\_location)  & (1) select\_army, (2) attack\_minimap (minimap\_location)\\
 Micros        & \textit{burrow\_lurkers}            & (1) select\_army, (2) select\_unit (lurker), (3) burrowdown\_lurker\_quick\\
 Scouting      & \textit{send\_scout} (minimap\_location)  & (1) select\_overlord, (2) move\_minimap (minimap\_location)\\
 \hline
\end{tabular}
\caption{Example macros available to each module. A macro (italic) consists of a sequence of macros or PySC2 actions (non-italic). Information such as base.minimap\_location, base.screen\_location and bases\_hotkey is provided by the updater. }
\label{table:macros}
\end{table*}


\section{Modular Architecture}

\begin{table}[!ht]
\centering
\begin{tabular}{|l|l|} 
 \hline
 Name & Description \\ [0.5ex] 
 \hline
 Time           & Total time (seconds) passed \\
 Friendly bases & Minimap locations and worker counts \\
 Enemy bases    & Minimap locations (scouted)\\
 Neutral bases  & Minimap locations \\
 Friendly units & Friendly unit types and counts \\
 Enemy units    & Enemy unit types and counts (scouted) \\
 Buildings      & All constructed building types  \\
 Upgrades       & All researched upgrades \\
 Build queue    & Units and buildings in production \\ 
 Notifications  & Any message from or to modules \\ [0.5ex] 
 \hline
\end{tabular}
\caption{Examples memories maintained by the updater}
\label{table:updater}
\end{table}

Table~\ref{table:modules} summarizes the role and design of each module. In the following sections, we will describe them in details for our implemented agent playing the Zerg race . Note that the design presented here is only an instance of all possible ways to implement this modular architecture. One can incorporate other methods, such as planning, into one of the modules as long as it works coherently with other modules.

\subsection{Updater}
The updater serves as a memory unit, a communication hub for modules, and a portal to the PySC2 environment. 

To allow a fair comparison between AI and humans, \namecite{vinyals_2017_starcraft} define observation inputs from PySC2 as similar to those exposed to human players, including imagery feature maps of the camera screen and the minimap (e.g. unit type, player identity), and a list of non-spatial features such as the total amount of minerals collected. Because past actions, past events, and out-of-camera information are crucial for decision making but not directly accessible from current observations, the agent has to develop an efficient memory. Though it is possible to learn such a memory from experiences, we think a properly hand-designed set of memories can serve a similar purpose, while also reducing the burden on reinforcement learning. Table ~\ref{table:updater} lists example memories the updater maintains. Some memories (e.g. build queue) can be inferred from previous actions taken. Some (e.g. friendly units) can be inferred from inspecting the list of all units. Others (e.g. enemy units) require further processing PySC2 observations and collaborating with the scouting module. 

The ``notifications'' entry holds any information a module wants to notify other modules, thus allowing them to communicate and cooperate. For example, when the build order module decides to build a new base, it notifies the tactics module, which may move armies to protect the new base. 

Finally, the updater handles communication between the agent and PySC2 by concretizing macros into sequences of PySC2 actions and executing them in the environment.

\subsection{Macros}
\noindent When playing StarCraft II, humans usually choose their actions from a list of subroutines, rather than from raw environment actions. For example, to build a new base, a player identifies an unoccupied neutral base, selects a worker, and then builds a base there. Here we name these subroutines as \textit{macros} (examples shown in Table~\ref{table:macros}). Learning a policy to output macros directly can hide the trivial execution details of certain higher level commands, therefore allowing the policy to explore different strategies more effectively.

\subsection{Build Order}
A StarCraft II agent must balance our consumption of resources between many needs, including supply (population capacity), economy, combat units, upgrades, etc. The build order module plays the crucial role of choosing the correct thing to build. For example, in the early game, the agent needs to focus on building enough workers to gather resources, and while in the mid game, it should choose the correct types of armies that can beat the opposing ones. Though there exist numerous efficient build orders developed by professional players, executing one naively without adaptation can result in highly exploitable behavior. Instead of relying on complex if-else logic or planning to handle various scenarios, the agent's build order module can benefit effectively from massive game-play experiences. Therefore we choose to optimize this module by deep reinforcement learning.

\begin{figure}[ht!]
    \centering
    \includegraphics[width=0.45\textwidth]{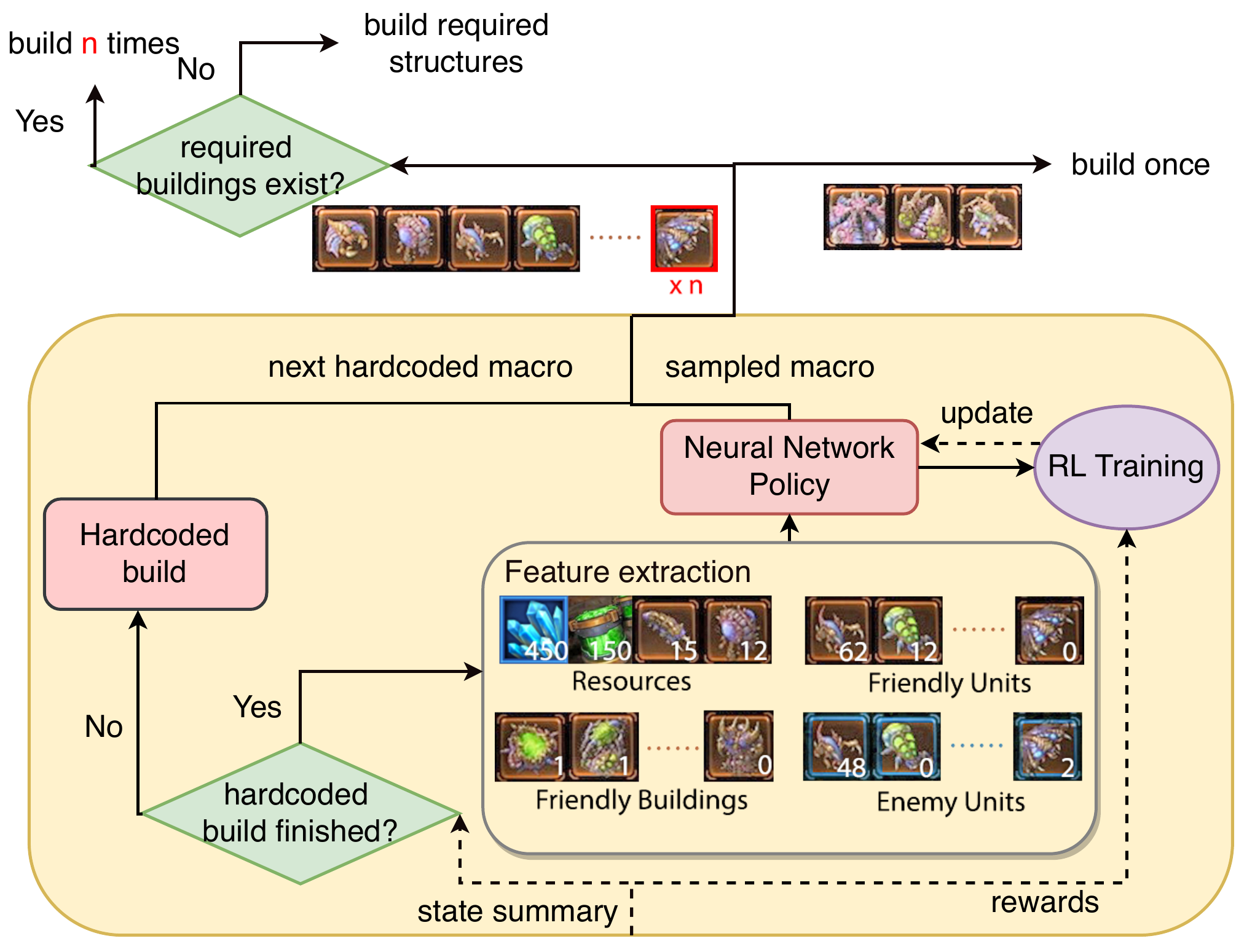}
    \caption{Details of our build order module.}
    \label{fig:build_order_diagram}
\end{figure}

Here we start with a classic hardcoded build order \footnote{Exactly the first 2 minutes taken from \\ https://lotv.spawningtool.com/build/56414/}, as the builds are often the same towards the beginning of the game, and the optimal trajectory is simple but requires precisely timed commands. Once the hard-coded build is exhausted, a neural network policy takes control (See Figure~\ref{fig:build_order_diagram}). This policy operates once every 5 seconds. Its input consists of the agent's resources (minerals, gas, larva, and supply), its building counts, its unit counts, and enemy unit counts (assuming no fog-of-war). We choose to exclude spatial inputs like screen and minimap features, because choosing what to build is a high-level strategic choice that depends more on the global information. The output is the type of unit or structure to produce. For units (Drones, Overlords, and combat units), it also chooses an amount $n \in \{1, 2, 4, 8, 16\}$ to build. For structures (Hatchery, Extractor) or Queen, it only produces one at a time. The policy uses a fully connected (FC) network with four hidden layers and 512 hidden units for each layer.

We also mask out invalid actions, such as producing more units than the current resources can afford, to enable efficient exploration. If a unit type requires a certain tech structure (e.g. Roaches need a Roach Warren) but it doesn't exist and is not under construction, then the policy will build the tech structure instead.

\subsection{Tactics}
Once our agent possesses an army provided by the build order module, it must learn to use it effectively. The tactics module handles map-level army commands, such as attacking or retreating to a specific location with a group of units. Though it is possible to hardcode certain tactics, we will show in the Evaluation section that a learned tactics can perform better.

In the current version, our tactics simply decides where on the minimap to move all of its armies towards. The input consists of $64 \times 64$ bitmaps of friendly units, enemy units (assuming no fog-of-war), and all selected friendly units on the minimap. The output is a probability distribution over minimap locations. The policy uses a three-layer Fully Convolutional Network (FCN) with 16, 32 and 1 filters, 5, 3 and 1 kernel sizes, and 2, 1 and 1 strides respectively. A softmax operation over the FCN output gives the probability over the minimap. The advantage of FCN is that its output is invariant to translations in the input, allowing the agent to generalize better to new scenarios or even new maps. The learned tactics policy operates every 10 seconds.  

\subsection{Scouting}
Because the fog-of-war hides certain areas, enemy-dependent decisions can be very difficult to make, such as building the correct army types to counter the opponent's.

Our current agent assumes no fog-of-war during self-play training, but can be evaluated under fog-of-war at test time, with a scouting module that supplies missing information. In particular, the scouting module sends Overlords to several predefined locations on the map, regularly moves the camera to those places and updates enemy information. It maintains an exponential moving average estimate of enemy unit counts for the build order module, and uses a neural network to predict enemy unit locations on the minimap for the tactics module. The prediction neural network applies two convolutions with 16, 32 filters and 5, 3 kernel sizes to the current minimap, followed by an LSTM of 512 hidden units, whose output is reshaped to the same size of the minimap, followed by pixel-wise sigmoid to predict the probabilities of enemy units. Future work will involve adding further predictions beyond enemy locations and using RL to manage scouts.

\subsection{Micromanagement} 
Micromanagement requires issuing precise commands to individual units, such as attacking specific opposing units, in order to maximize combat outcomes. Here we use simple scripted micros in our current version, leaving the learning of more complex behavior to future work, for example, by leveraging existing techniques presented in the Related Work. Our current micromanagement module operates when the updater detects that friendly units are close to enemies. It moves the camera to the combat location, groups up the army, attacks the location with most enemies nearby, and uses a few special abilities (e.g. burrowing lurkers, spawning infested terrans).

\subsection{Worker Management}
Worker management has been extensively studied for StarCraft: Brood War \cite{christensen_2010_resource}. StarCraft II simplifies the process, so the suggested worker assignment (2 per mineral patch, 3 per vespene geyser) is often sufficient for professional players. We script this module by letting it constantly review worker counts of all bases and transferring excess workers to under-saturated mining locations, prioritizing gas over minerals. It also periodically commands Queens to inject larva at all bases, which is crucial for boosting unit production.

\subsection{Scheduler}
The PySC2 environment places a restriction on the number of actions per minute (APM) to ensure a fair comparison between AI and human. Therefore when multiple modules propose too many macros at the same time, not all macros can be executed and a scheduler is required to order them by priority. Our current version uses little APM, so the scheduler simply cycles through all modules and executes the oldest macro proposals. When APM increases in the future, for example when complex micromanagement comes into play, a cleverer or even learned scheduler will be required.

\section{Training Procedure}
Our agent is trained to play Zerg v.s. Zerg on the ladder map Abyssal Reef on the 4.0 version of StarCraft II. For most games, Fog-of-war is disabled. Note that built-in bots can also utilize full observations, so the comparison is fair. Each game lasts 60 minutes, after which a tie is declared. 

\subsection{Self-Play}
We follow the self-play procedure suggested by \namecite{bansal_2018_emergent} to  save snapshots of the current agent into a training pool periodically (every $3 \times 10^6$ policy steps). Each game the agent plays against a random opponent sampled uniformly from the training pool. To increase the diversity of opponents, we initialize the training pool with a random agent and other scripted modular agents that use fixed build orders and simple scripted tactics. The fixed build orders are optimized\footnote{Most builds taken from https://lotv.spawningtool.com/build/zvz/} to prioritize specific unique types and include \textit{zerglings}, \textit{banelings}, \textit{roaches\_and\_ravagers}, \textit{hydralisks}, \textit{mutalisks}, \textit{roaches\_and\_infestors}, and \textit{corruptors\_and\_broodlords}. Zerglings are available to every build order. The scripted tactics attacks the enemy bases with all armies whenever its army supply is above 50 or its total supply is above 100. The agent never faces the built-in bots until test time.

\subsection{Reinforcement Learning}
If winning games is the only concern, in principle the agent should only receive a binary win-loss reward. However, we have found that the win-loss provides too sparse training signals and thus slows down training. Instead we use the supply difference ($d_t$) between the agent and the enemy as a reward function. A positive supply difference is often correlated with an advantageous status. Specifically, to ensure that the game is always zero-sum, the reward is the \textit{change} in supply difference $r_t = d_t - d_{t-1}$ for each time step. Summing up all rewards yields a total reward equal to the end-game supply difference.

We use Asynchronous Advantage Actor-Critic \cite{mnih_2016_asynchronous} to optimize the policies with 18 parallel CPU workers. The learning rate is $10^{-4}$ and the entropy bonus coefficient is $10^{-1}$ for build order, $10^{-4}$ for tactics (smaller due to a larger action space). Each worker commits a gradient update to the central parameter server every 3 minutes in game (every 40 gradient steps for build order and 20 for tactics)

\subsection{Iterative Training}
One major benefit of the modular architecture is that modules act relatively independently and can therefore be optimized separately. We illustrate this by comparing iterative training, namely optimizing one module while keeping others fixed, against joint training, namely optimizing all modules together. We hypothesize that iterative training can be more effective because it stabilizes the experiences gathered by the training module and avoids the complex module-wise coordination during joint training. 

In particular, we pretrain a build order module with a scripted tactics described in the Self Play section, and meanwhile pretrain a tactics module with a scripted build order (all Roaches). Once both pretrained modules stabilize, we combine them, freeze the tactics, and only train the build order. After build order stabilizes, we freeze its parameters and train tactics instead. The procedure is abbreviated ``iterative, pretrained build order and tactics''. 
\section{Evaluation \label{sec:evaluation}}
Videos our agent playing against itself and qualitative analysis of the tactics module are available on https://sites.google.com/view/modular-sc2-deeprl. In this section, we would like to analyze the quantitative and qualitative performance of our agent by answering the following questions.

\begin{enumerate}
    \item Does our agent trained with self-play outperform scripted modular agents and built-in bots?
    \item Does iterative training outperform joint training?
    \item How does the learned build order behave qualitatively? E.g. does it choose army types that beat the opponent's? 
    \item Can our agent generalize to other maps after being trained on only one map?
\end{enumerate}

\subsection{Quantitative Performance}

\begin{figure}[!ht]
    \centering
    \includegraphics[width=0.45\textwidth]{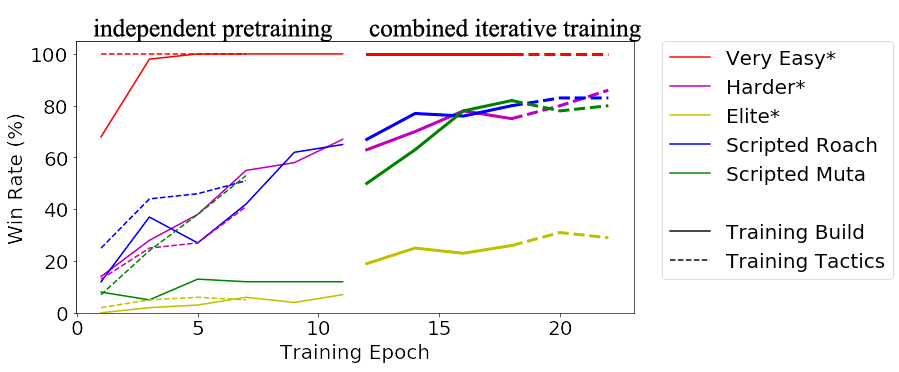}
    \caption{Win rates of our agent against opponents of different strengths. Asterisks indicate built-in bots that are not seen during training. $1$ epoch = $3 \times 10^5$ policy steps.}
    \label{fig:self_play_winrate}
\end{figure}

Figure~\ref{fig:self_play_winrate} shows the win rates of our agent throughout training, under the ``iterative, pretrained build order and tactics'' procedure. Pretrained build order and tactics can already achieve 67\% and 41\% win rates against the Harder bot. The win rate of combined modules increase to 87\% after iterative training. Moreover, it outperforms simple scripted agents (also see Table~\ref{table:winrates_vs_bots}), indicating the effectiveness of reinforcement learning.

Table~\ref{table:training_procedures} shows that iterative training outperforms joint training by a large margin. Even when joint training also starts with a pretrained bulid order, its stability quickly drops and results in 52\% less win rate than iterative against the Harder bot. Pretraining both build order and tactics lead to better overall performance.

\subsection{Qualitative Evaluation of the Learned Build Order}
\begin{figure}[!ht]
    \centering  
    \includegraphics[width=0.45\textwidth]{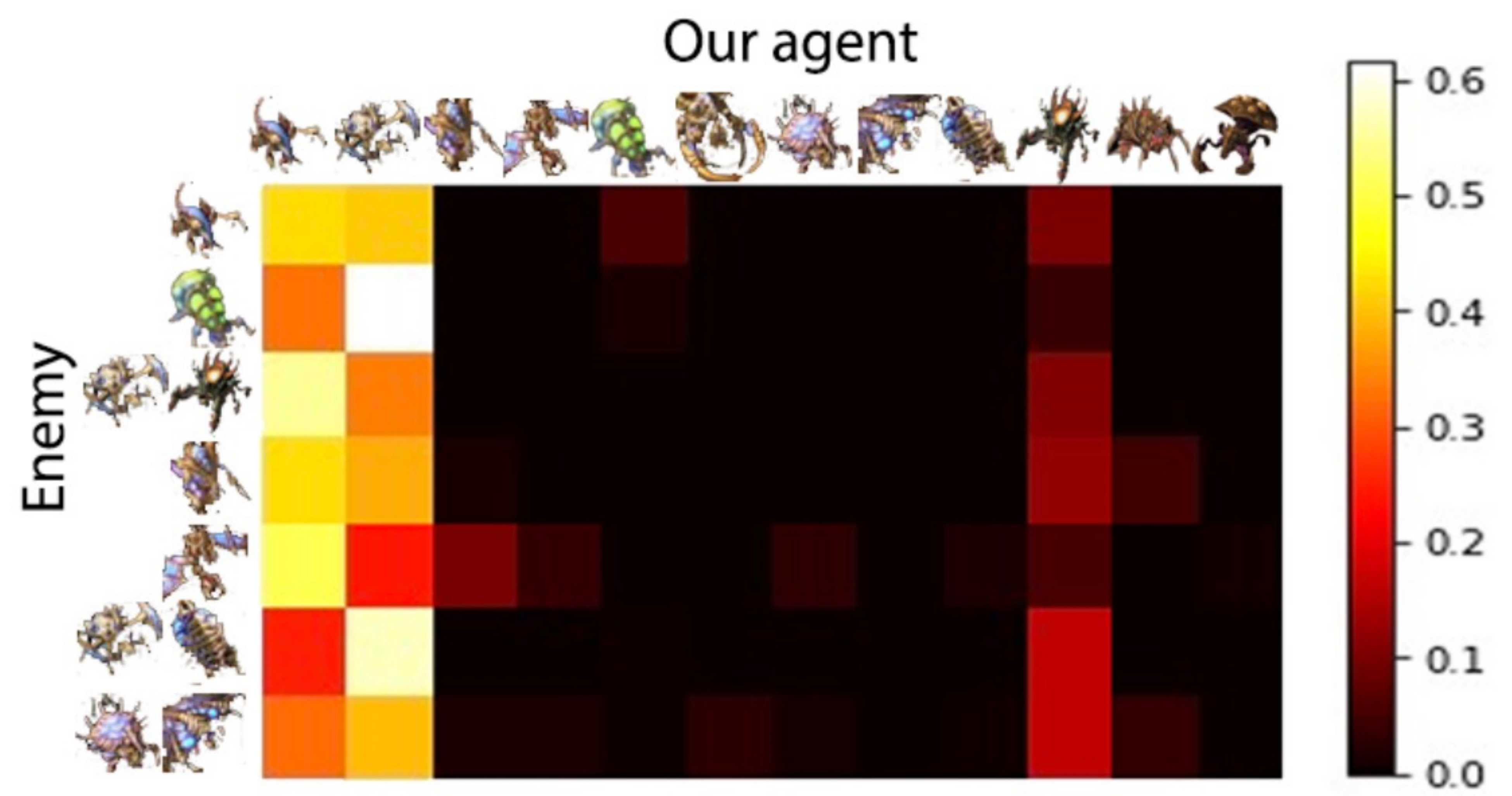}
    \caption{Learned army compositions. Showing ratios of total productions of each unit type to the total number of produced combat units.}
    \label{fig:army_composition}
\end{figure}

\begin{table*}[t]  
\centering
\begin{tabular}{|l||l|l|l|l|} 
 \hline
 Training procedure                 & Hard & Harder & Very Hard & Elite \\ [0.5ex] 
 \hline
 Iterative, no pretrain             & 77\pmm 6 \%   & 62\pmm 10\%  & 11\pmm 7\% & 11\pmm 5\%\\
 Iterative, pretrained tactics      & 84\pmm 4 \%   & 73\pmm 8\%  & 16\pmm 5\% &  9\pmm 6\% \\
 Iterative, pretrained build order  & \textbf{85\pmm 7\%}   & 81\pmm 5\%  & 39\pmm 9\% & 25\pmm 8\% \\
 Iterative, pretrained build order and tactics & 84\pmm 5\% & \textbf{87\pmm 3\%} & \textbf{57\pmm 7\%} & \textbf{31\pmm 11\%} \\
 Joint, no pretrain                 & 41\pmm 21\%   & 21\pmm 13\%  & 4\pmm 4\%  &  5\pmm 4\% \\
 Joint, pretrained build order      & 65\pmm 18\%   & 35\pmm 9\%  & 10\pmm 4\%  & 6\pmm 3\% \\
 \hline
\end{tabular}
\caption{Comparison of final win rates between different training procedures against built-in bots (3 seeds, averaged over 100 matches per seed)}
\label{table:training_procedures}
\end{table*}

\begin{table*}[t]  
\centering
\begin{tabular}{|l||l|l|l|l|l|l|l|l|l|l|} 
 \hline
 & Average & Roach & Mutalisk & V. Easy & Easy & Medium & Hard & Harder & V. Hard & Elite \\ [0.5ex] 
 \hline 
 Modular (None) & 67\pmm 2\% & 69\pmm 3\% & 41\pmm 6\% & 100\% & 92\pmm 2\% & 71\pmm 4\% & 77\pmm 3\% & 62\pmm 9\% & 11\pmm 4\% & 11\pmm 5\%\\
 Modular (Tactics) & 63\pmm 4\% & 76\pmm 6\% & 29\pmm 5\% & 100\% & 100\% & 91\pmm 2\% & 84\pmm 3\% & 73\pmm 7\% & 16\pmm 6\% & 9\pmm 4\%\\
 Modular (Build) & 76\pmm 3\% & 81\pmm 3\% & 43\pmm 6\% & 100\% & 96\pmm 1\% & 92\pmm 1\% & \textbf{85\pmm 2\%} & 81\pmm 3\% & 39\pmm 7\% & 25\pmm 5\%\\
 Modular (Both) & \textbf{83\pmm 2\%} & 80\pmm 7\% & \textbf{67\pmm 5\%} & \textbf{100\%} & \textbf{100\%} & \textbf{99\%} & 84\pmm 3\% & \textbf{87\pmm 2\%} & \textbf{57\pmm 5\%} & \textbf{31\pmm 10\%} \\
 Scripted Roaches & 62\pmm 3\% & -- & 18\pmm 9\% & 100\% & 100\% & 92\pmm 3\% & 71\pmm 7\% & 35\pmm 4\% & 6\pmm 3\% & 5 \pmm 2\%\\
 Scripted Mutalisks & 71\pmm 2\% & \textbf{82\pmm 5\%} & -- & 100\% & 99\% & 86\pmm 4\% & 77\pmm 5\% & 64\pmm 7\% & 15\pmm 4\% & 2 \pmm 1\%\\
 \hline
\end{tabular}
\caption{Comparison of win rates (out of 100 matches) against various opponents. Pretrained component in parenthesis. ``V.'' means ``Very''.}
\label{table:winrates_vs_bots}
\end{table*}

Figure~\ref{fig:army_composition} shows how our learned build order module reacts to different scripted opponents that it sees during training. Though our agent prefers the Zergling-Roach-Ravager composition in general, it can correctly react to the Zerglings with more Banelings, to Banelings with fewer Zerglings and more Roaches, and to Mutalisks with more hydralisks. Currently, the reactions are not perfectly tailored to the specific opponents, likely because the Zergling-Roach-Ravager composition is strong enough to defeat the opponents before they can produce enough units.

\subsection{Generalization to Different Maps}
Many parts of the agent, including the modular architecture, macros, choice of policy inputs, and neural network architectures (specifically FCN), are designed with certain prior knowledge that can help with generalization to different scenarios. We test the effect of prior knowledge by evaluating our agent against different opponents on maps not seen during training. These test maps have various sizes, terrains, and mining locations. The 4-player map Darkness Sanctuary even randomly spawns players at 2 out or 4 locations. Table~\ref{table:map_generalization} summarizes the results. Though our agent's win rates drop by 7.5\% on average against Harder, it is still very competitive.

\begin{table}[h!]
\centering
\begin{tabular}{|l|l|l|l|} 
 \hline
 Opponent           & AR    & DS    & AP \\ [0.5ex]
 \hline
 Scripted Roaches   & 80\pmm 7\%  & 82\pmm 4\%  & 78\pmm 11\% \\ 
 Hard               & 84\pmm 3\%  & 84\pmm 6\%  & 78\pmm 5\% \\ 
 Harder             & 87\pmm 2\%  & 77\pmm 7\%  & 82\pmm 6\% \\ 
 Very Hard          & 57\pmm 5 \%  & 44\pmm 7\%  & 55\pmm 4\% \\ 
 Elite              & 31\pmm 10 \%  & 22\pmm 11\%  & 30\pmm 10\% \\ 
 \hline
\end{tabular}
\caption{Win rates  (out of 100 matches) of our agent against different opponents on various maps. Our agent is only trained on AR. AR = Abyssal Reef. DS = Darkness Sanctuary. AP = Acid Plant.}
\label{table:map_generalization}
\end{table}

\begin{table}[h!]
\centering
\begin{tabular}{|l|l|l|l|l|} 
 \hline
 Opponent           & Hard & Harder & V. Hard & Elite \\ [0.5ex]
 \hline
 Win Rate   & 95\pmm 1\%  & 94\pmm 2\%  & 50\pmm 8\%  & 60\pmm 8\%\\ 
 \hline
\end{tabular}
\caption{Win rates  (out of 100 matches) of our agent on Abyssal Reef with fog-of-war enabled}
\label{table:fog}
\end{table}

\subsection{Testing under Fog-of-War}
Though the agent was trained without fog-of-war, we can test its performance by filling missing information with estimates from the scouting module. Table~\ref{table:fog} shows that the agent actually performs much better under fog-of-war, achieving 10\% higher win rates on average, potentially because the learned build orders and tactics generalize better to noisy/imperfect information, while the built-in agents rely on concrete observations. 

\section{Conclusions and Future Work}
In this paper, we demonstrate that proper combination of human knowledge and deep reinforcement learning can result in competitive StarCraft II agents. Certain techniques like modular architecture, macros, and iterative training can provide insights to dealing with other challenging problems at the scale of StarCraft II.

However, our current version is still far from beating the hardest built-in bots, let alone skilled humans. Many improvements are under research, including deeper neural networks, multi-army-group tactics, researching upgrades, and learned micromanagement policies. We believe that such improvements can eventually close the gap between our modular agent and professional human players.

\section{Acknowledgements}
This work was supported in part by the DARPA XAI program and Berkeley DeepDrive.

\bibliographystyle{aaai}
\bibliography{ref}

\end{document}